\documentclass[letterpaper, 10 pt, conference]{ieeeconf}  

\IEEEoverridecommandlockouts                              

\usepackage{graphics} 
\usepackage{epsfig} 
\usepackage{mathptmx} 
\usepackage{times} 
\usepackage{amsmath} 
\usepackage{amssymb}  
\usepackage{url}
\usepackage{bm}
\usepackage{float}

\title{\LARGE \bf
Load-Based Variable Transmission Mechanism for Robotic Applications
}

\author{Sinan Emre$^{1,2}$, Victor Barasuol$^{1}$, Matteo Villa$^{1}$ and Claudio Semini$^{1}$ 
\thanks{$^{1}$ Dynamic Legged Systems (DLS) lab, Istituto Italiano di Tecnologia
(IIT), Genova (Italy). E-mail: 
        {\tt\small name.surname@iit.it}. Research partially funded by the European Union - NextGenerationEU and by the Ministry of University and Research (MUR), National Recovery and Resilience Plan (NRRP), Mission 4, Component 2, Investment 1.5, project “RAISE - Robotics and AI for Socio-economic Empowerment” (ECS00000035)}%
\thanks{$^{2}$  Dipartimento di Informatica, Bioingegneria, Robotica e Ingegneria dei
Sistemi (DIBRIS), University of Genoa, Genoa, Italy.}%
}

\begin{document}

\maketitle
\thispagestyle{empty}
\pagestyle{empty}

\begin{abstract}

This paper presents a Load-Based Variable Transmission (LBVT) mechanism designed to enhance robotic actuation by dynamically adjusting the transmission ratio in response to external torque demands. Unlike existing variable transmission systems that require additional actuators for active control, the proposed LBVT mechanism leverages a pre-tensioned spring and four-bar linkage to passively modify the transmission ratio, thereby reducing the complexity of the robot joint actuation system. To evaluate the effectiveness of the LBVT mechanism, we conducted simulation-based analyses. The results confirm that the system achieves a 40$\%$ increase in transmission ratio upon reaching a predefined torque threshold, effectively amplifying joint torque when needed, without additional actuation. Furthermore, the simulations demonstrate a torque amplification effect, triggered when the applied force exceeds 18N, highlighting the system’s ability to autonomously respond to varying load conditions. This research contributes to the development of lightweight, efficient, and adaptive transmission systems for robotic applications, particularly in legged robots, where dynamic torque adaptation is crucial. 

\end{abstract}

\section{Introduction}

Variable transmission mechanisms (VTMs) are crucial in robotic applications as they enable adaptive control of torque and speed, enhancing both energy efficiency and mechanical performance. Traditional fixed-gear systems and constant reduction-ratio actuators limit the adaptability of robotic systems, especially for legged robots and in dynamic environments where load conditions frequently change. These constraints often result in increased energy consumption, suboptimal torque utilization, and mechanical inefficiencies.

VTM technology has been developed for several decades. Hirose et al. \cite{hirose1999development} introduced the X-Screw, a load-sensitive actuator featuring a variable transmission mechanism that automatically adjusts to external load conditions. The X-Screw actuator operates by using an eccentric nut-screw mechanism where the nut's offset changes dynamically based on the applied axial load. Further advancements were made with radial crank-type CVTs (Continuously Variable Transmission) that use two ball screws to apply controlled forces to radially connected links, which drive a crank through a linear guide mechanism \cite{yamada2012radial}. By adjusting the crank length and angle, the system continuously varies the transmission ratio, allowing efficient torque and speed control while overcoming the motion range limitations of traditional crank-type CVTs. 

More studies have focused on actuator-based designs to enhance adaptability and performance. Dual-motor (one optimized for high torque and the other for high speed) planetary gear systems were developed to improve actuation efficiency by distributing power across multiple transmission pathways \cite{lee2011new}.  Kim et al. \cite{kim2007double} introduced a double actuator unit (DAU) incorporating a planetary gear train, enabling simultaneous position and stiffness control, thereby improving force control and collision safety in robotic manipulators. Another study presented a robotic system that dynamically adjusts its actuator gear-ratio to amplify or mitigate natural load dynamics \cite{girard2015two} \cite{girard2017leveraging}.

VTMs have also been explored in robotics applications, such as a proposed jumping rescue robot that utilized a VTM to adjust jumping height and distance dynamically \cite{dunwen2011concept}, demonstrating the potential of these applications in robotic systems. They have also been implemented in robotic prostheses, such as the SuKnee \cite{sun2018variable}, which uses a slider-crank mechanism to dynamically adjust transmission ratios based on knee angles, enhancing both torque output for standing and speed variation for walking, demonstrating their effectiveness in wearable robotics.

Expanding on earlier research, recent studies have refined VTMs by enhancing their dynamic adaptability, improving energy efficiency, and enabling multi-modal functionality across various robotic applications. A notable example is the further development of the fast gear-shifting Dual-Speed Dual-Motor (DSDM) actuator \cite{girard2015two}, which now enables transitions between two distinct reduction ratios, enhancing efficiency and adaptability for robotic tasks involving dynamic contact situations \cite{girard2022fast}. 

At the actuator level, another approach was introduced by Meng et al. \cite{meng2022explosive}, who demonstrated the viability of explosive electric actuators incorporating a two-stage planetary reducer transmission, effectively implementing them in both quadruped and humanoid robots for high-power jumping tasks. The Dual Reduction Ratio Planetary Drive (DRPD) actuator enhances robotic actuation by switching between low and high reduction ratios using a 3K compound planetary drive and a pawl brake mechanism, optimizing both speed and torque for various robotic tasks \cite{song2022drpd}. Similarly, the Variable Transmission Series Elastic Actuator (VTSEA) for hip exoskeletons dynamically adjusts its transmission ratio across three discrete levels, improving torque-speed adaptability and enhancing transparency in human-robot interactions \cite{wang2024development}. These advancements demonstrate how variable transmission technologies are integrated into robotic and wearable systems to enhance efficiency, adaptability, and performance across diverse applications. 

There are also interesting approaches like the Elastomeric Continuously Variable Transmission (ElaCVT), which passively adjusts its reduction ratio based on external loads through elastomer deformation \cite{kim2020elastomeric} \cite{shin2022passively}. The applicability of the ElaCVT  is constrained by structural limitations, including insufficient material rigidity, low transmission efficiency, restricted reduction ratio adjustment, and friction-induced slippage. Similarly, the Mechanical Variable Magnetic Gear Transmission (MVMGT) employs a non-contact coaxial magnetic gear system, where the transmission ratio is modified by mechanically adjusting the magnetic harmonics between the rotors \cite{lee2022mechanical}.

Building on these advancements in transmission mechanisms, Nobaveh et al. proposed the Compliant Continuously Variable Transmission (CVT) that utilizes the warping of twisting beams, where a central rotational constraint enables continuous adjustment of the transmission ratio based on beam deformation \cite{nobaveh2023compliant}. Likewise, Kim et al. \cite{kim2020optimization} proposed the bio-inspired variable gearing system that replicates the behaviour of pennate muscles, dynamically modifying the transmission ratio to balance speed and force in robotic actuators. Furthermore, Sanz et al.  \cite{sanz2023active} proposed an active knee orthosis featuring a variable transmission ratio that utilizes a dual-clutch mechanism to alternate between high-torque and high-speed modes efficiently, enhancing adaptability in wearable robotics for activities like walking and sit-to-stand movements. Barry, a high-payload quadruped robot, served as the basis for an actively variable transmission (AVT) robotic leg, where researchers implemented and tested a variable transmission mechanism in the leg structure \cite{valsecchi2023barry} \cite{valsecchi2023actively}. This AVT system dynamically adjusts its transmission ratio to optimize power efficiency and agility, demonstrating a reduction of knee actuator power consumption by up to 50\% through controlled experiments on a single-leg prototype. Shin et al. \cite{shin2025variable} developed a Variable Transmission with Actively Controllable Reduction Ratio, utilizing a toroidal drive and a compression-based power transmission system that actively adjusts the gear ratio while maintaining high efficiency and built-in overload protection. Meanwhile, Hur et al. \cite{hur2024continuously} proposed the Continuously Variable Transmission and Variable Stiffness Actuator (CVT-VSA) that utilizes an actively adjustable four-bar linkage, enabling real-time modulation of both transmission ratio and stiffness to enhance torque-speed adaptability and mitigate impacts in highly dynamic robotic systems.

These approaches can be widely categorized into two groups. The first includes active VTMs, which achieve transmission shifts through an additional actuator but at the cost of increased complexity, weight, and energy consumption. The second group corresponds to passive VTMs, which avoid the need for a secondary actuator yet often show limited adaptability to rapid or unpredictable load variations. To overcome these limitations, this study introduces a Load-Based Variable Transmission (LBVT) mechanism that dynamically responds to external torque demands while preserving the simplicity of a fully passive design. Our contributions include:
\begin{itemize}

    \item A novel passive LBVT mechanism that reacts instantaneously to torque demand by leveraging a passive adaptation approach. It utilizes pre-tensioned springs and a four-bar linkage to dynamically modify the transmission ratio in response to external forces.
    
    \item The proposed mechanism eliminates the need for extra actuators to shift transmission ratios, reducing mechanical complexity and system weight. By leveraging passive dynamics, it maintains a lightweight design while improving torque delivery under varying load conditions.

\end{itemize}

\section{Mechanical Design and Transmission Characteristics}

\subsection{Fundamental Concept of the Mechanism}

In this section, we introduce the concept of the proposed LBVT mechanism. Even though developed for the joints of legged robots, the mechanism can also be applied to other robot assemblies, such as manipulator arms. 

\begin{figure} [h]
    \centering
    \includegraphics[width=0.8\linewidth]{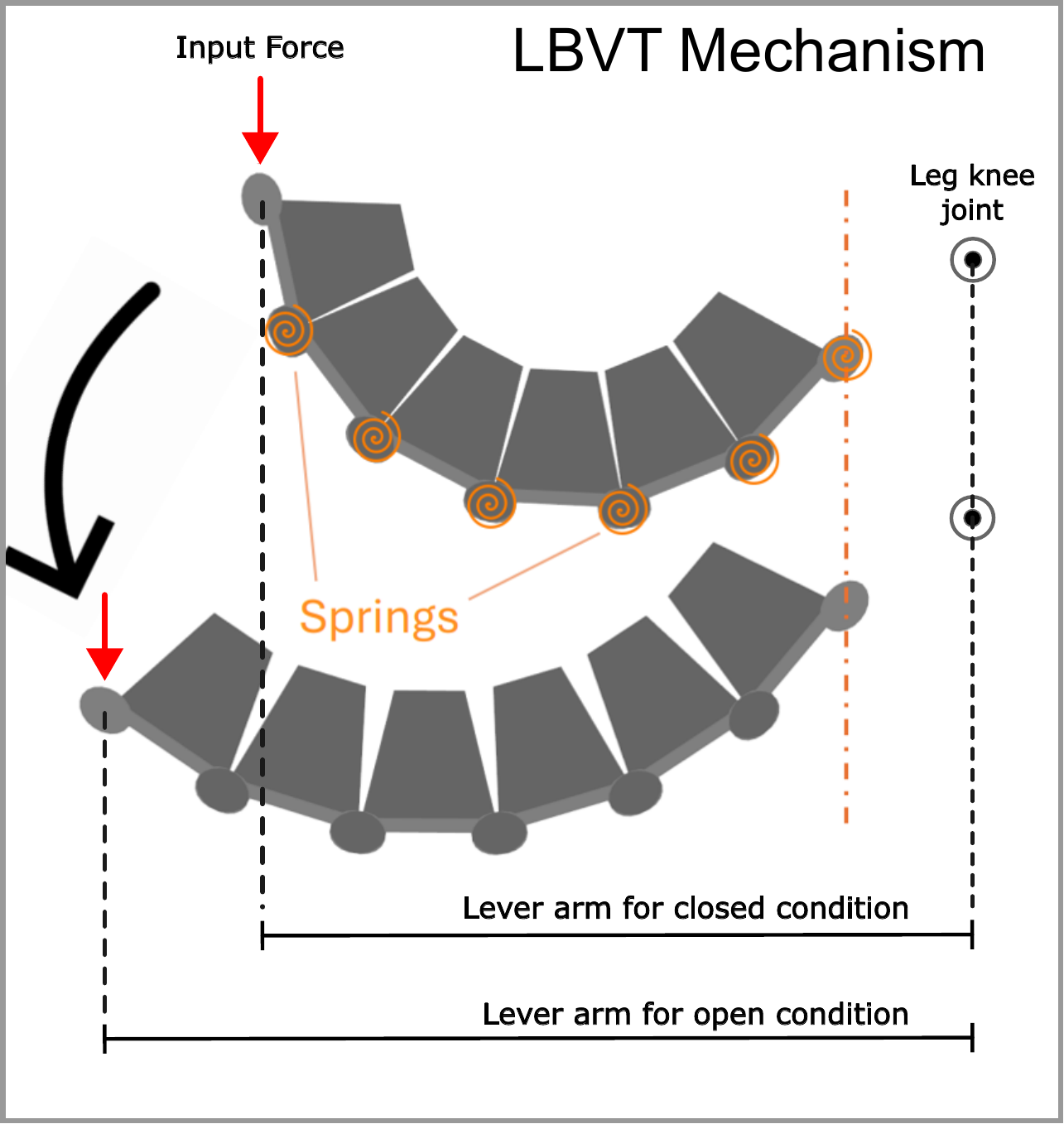}
    \caption{Concept drawing of the proposed LBVT mechanism. The closed mechanism starts to open once an external force exceeds a threshold force. The LBVT mechanism extends under the external force until it reaches mechanical limits.}
    \label{Concept}
\end{figure}

Fig. \ref{Concept} depicts a concept drawing of the proposed LBVT mechanism. This mechanism consists of multiple interconnected links that share a uniform structural configuration. The links are strategically arranged in series, with each segment linked by a pre-tensioned spring.

A key feature of this mechanism is its ability to respond dynamically to applied torque. The pre-tensioned spring system starts to extend once the knee joint experiences a torque that surpasses a predefined threshold. We defined this pre-tension mechanism to ensure that, under loads below the threshold, the mechanism remains inactive, and the transmission change only occurs after a certain limit is exceeded. This load-based activation leads to the extension of the lever arm and thus increases the transmission ratio. The increased lever length increases the mechanism's overall torque output. By leveraging this controlled extension, the system varies its transmission ratio when needed, thereby improving its mechanical performance.

\begin{figure} [h]
    \centering
    \includegraphics[width=0.9\linewidth]{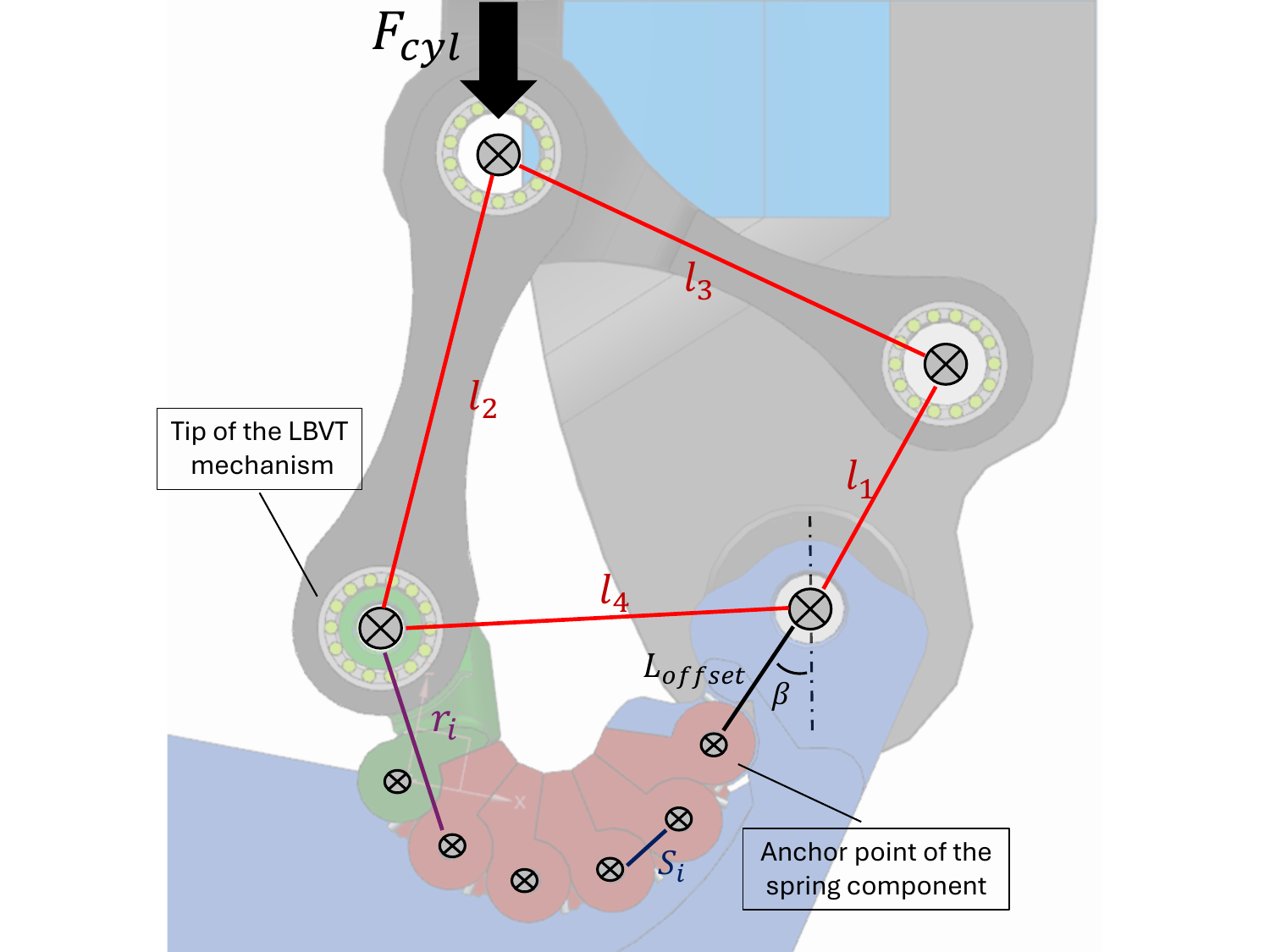}
    \caption{CAD drawing of the knee joint mechanism highlighting the four-bar linkage and its four links $l_1$-$l_4$. $l_1$ is the distance between the KFE (Knee Flexion Extension) joint and the four-bar linkage. $l_2$ and $l_3$ are the lengths of the bars. $l_4$ is the distance between the KFE joint and the tip of the mechanism.}
    \label{Kinmeatics}
\end{figure}

\subsection{Kinematic and Dynamic Modelling}

 The LBVT mechanism consists of a four-bar linkage with a pre-tensioned spring that dynamically adjusts the transmission ratio based on the external torque demand. The four-bar linkage shown in Fig. \ref{Kinmeatics} has a fixed frame ($l_1$), input link ($l_2$), coupler link ($l_3$), and output link ($l_4$). Additionally, the kinematic behaviour of the four-bar linkage between the actuator and the knee joint, excluding the influence of the spring, was previously studied in the context of the HyQ2Max quadruped robot \cite{semini2016design}. In this study, the relationship between the KFE joint torque $T(\theta)$ and the linear actuator force $F_{cyl}$ is shown as (\ref{fourbar_kine}), where $\theta$ is the KFE joint angle and $J(\theta)$ is the Jacobian.

 \begin{equation}
     T(\theta) = J^T(\theta)\cdot F_{cyl}
     \label{fourbar_kine}
 \end{equation}

 With the incorporation of the LBVT mechanism, the parameter $l_4$, which defines the distance between the tip of the mechanism and the KFE joint, becomes a variable parameter. Therefore, the KFE joint torque $T(\theta)$ is now dependent on this variability, leading to a modified torque expression at (\ref{spring_kine}). 

\begin{equation}
     T(\theta,l_4) = J^T(\theta,l_4)\cdot F_{cyl}
     \label{spring_kine}
 \end{equation}

\begin{equation}
    l_4 = \sqrt{x_{l_4}^2 + y_{l_4}^2}
    \label{$l_4$ kine1}
\end{equation}

\begin{equation}
    \begin{aligned}
        x_{l_4} = L_{offset}cos(\beta) +  \sum_{i = 1}^6{S_icos(\beta + \sum_{k = 1}^i{\alpha_k})}\\
        y_{l_4} = L_{offset}sin(\beta) +  \sum_{i = 1}^6{S_isin(\beta + \sum_{k = 1}^i{\alpha_k})}
    \end{aligned}
    \label{$l_4$ kine2}
\end{equation}

The \textbf{transmission ratio} varies dynamically as $l_4$ changes between its minimum and maximum limits. This relationship is derived using  (\ref{$l_4$ kine1}) and (\ref{$l_4$ kine2}), where $L_{offset}$ denotes the distance between the knee joint and the anchor point of the spring, and $\beta$ defines the orientation angle of the spring with respect to the vertical axis. $S_i$ indicates the segment length of each spring element, while $\alpha_k$ represents the deflection angle at the $k^{th}$ joint of the spring chain.  

Since $l_4$ directly influences the mechanical leverage of the four-bar linkage, its variation alters the torque amplification characteristics, leading to \textbf{variable transmission ratio} throughout the motion.

As previously discussed, each joint of the spring element is pre-tensioned, imparting the mechanism with its characteristic triggering behaviour. The fundamental effect of this pre-tension on a single joint is expressed in (\ref{preload-1}) where  $k_{joint_i}$ denotes the torsional stiffness of the 
$i^{th}$ joint and $T_{S_i}$ represents the corresponding joint torque. Given that the force at the tip of the mechanism can be determined from (\ref{preload-2}), the resulting torque at each spring joint can be systematically mapped.

\begin{equation}
    |T_{S_i}|\ge |k_{joint_i} * \alpha_{preload}| 
    \label{preload-1}
\end{equation}

\begin{equation}
     T(\theta,l_4) = F_{end}\cdot l_4
    \label{preload-2}
\end{equation}

Then, the torque at each joint of the spring chain can be calculated by (\ref{preload-3}), where $\gamma = \phi + \alpha + \theta_{l_4}$. Here, $\phi$ denotes the fixed offset angle for each spring joint, $\theta_{l_4}$ represents the absolute orientation of the $l_4$ link relative to the last segment of the spring.

\begin{equation}
    T_{s_i} = \sum_{i = 1}^6 r_i\cdot sin(\gamma_i)\cdot F_{end}
    \label{preload-3}
\end{equation}

\begin{equation}
    min|\sum_{i = 1}^6 r_i\cdot sin(\gamma_i)\cdot \frac{J^T(\theta,l_4)\cdot F_{cyl}}{l_4}| \ge |k_{joint_i} * \alpha_{preload}|
    \label{preload-4}
\end{equation}

Since triggering occurs only when the torque at one of the joints reaches the minimum threshold for motion, the relationship between the triggering effect and the input force is expressed by (\ref{preload-4}). The kinematic and dynamic model was developed under specific assumptions to simplify the analysis. Friction, backlash, and link compliance were not considered, and all components were treated as rigid bodies. These assumptions make it possible to isolate the effect of the pre-tension behaviour and the transmission characteristics of the mechanism.

\subsection{Mechanical Design}

\begin{figure} [ht]
    \centering
    \includegraphics[width=1\linewidth]{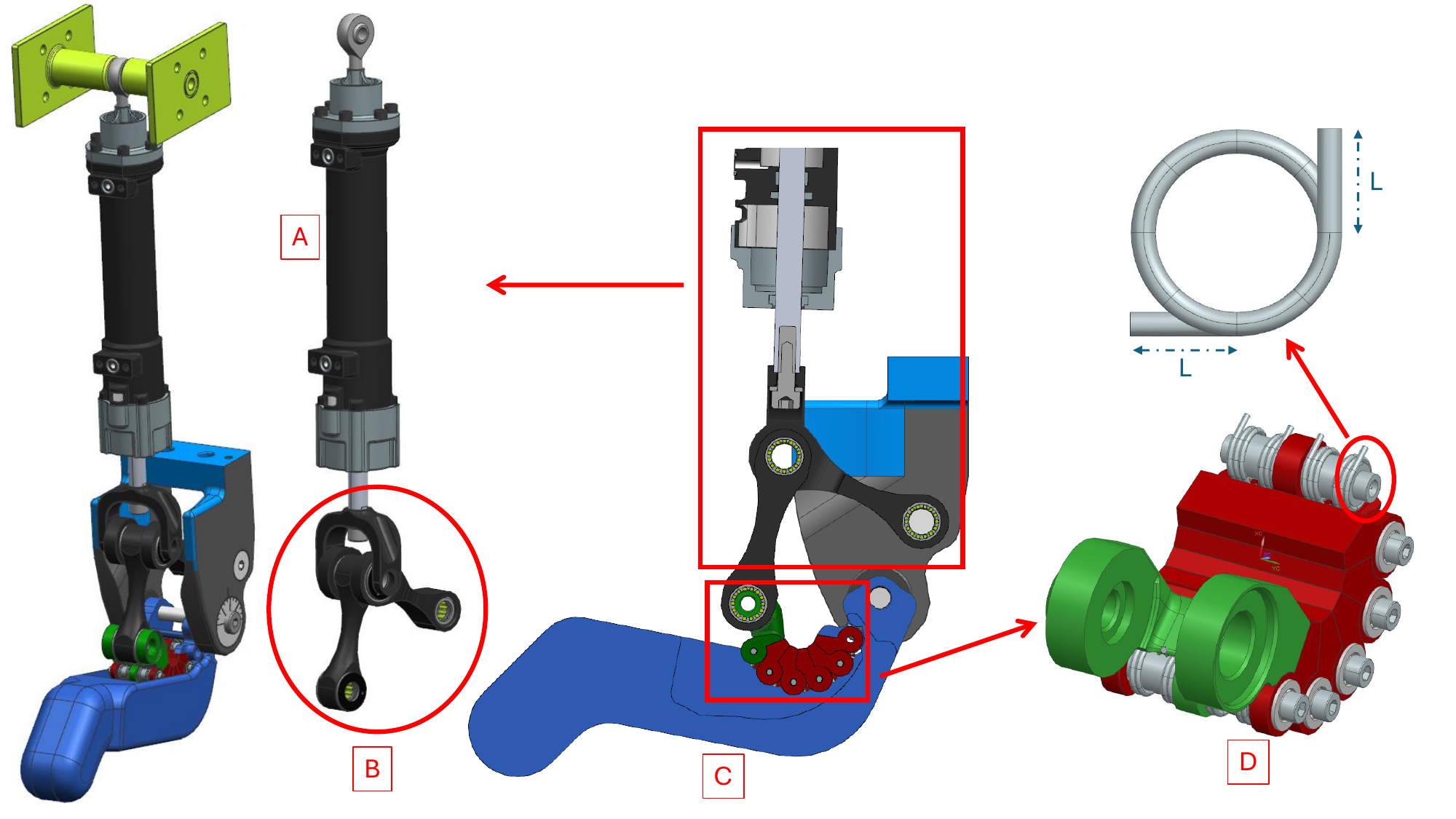}
    \caption{CAD view of the LBVT mechanism. The LBVT mechanism consists of a four-bar linkage integrated with a spring element and an end stop embedded within the lower leg. A is the linear actuator that is connected to the B four-bar mechanism. C shows the lower leg design with the full LBVT mechanism, and D represents the spring component of the LBVT mechanism.}
    \label{CAD}
\end{figure}

\begin{figure} [ht]
    \centering
    \includegraphics[width=1\linewidth]{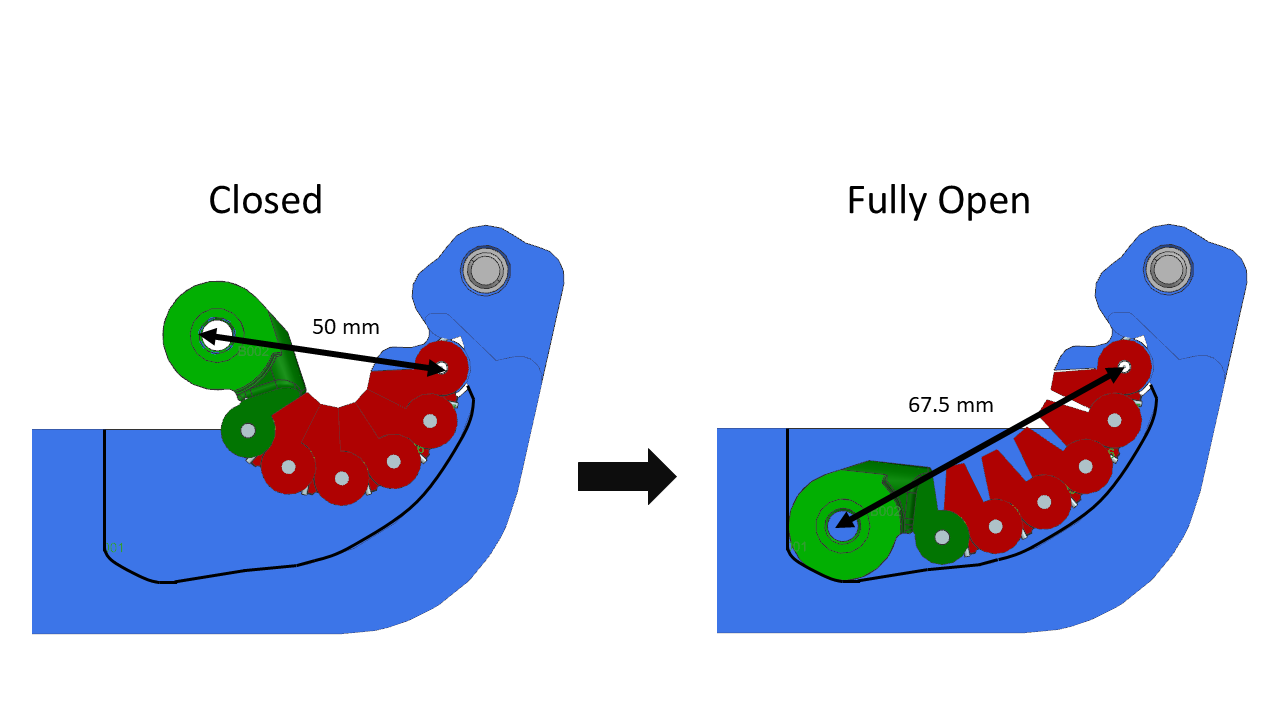}
    \caption{Extension of the LBVT mechanism and the lower leg design, which incorporates an internal structure that acts as a mechanical end-stop for the LBVT mechanism. On the left, the mechanism is closed, and on the right, the mechanism is fully open.}
    \label{Spring Opening}
\end{figure}

 Fig. \ref{CAD} shows the computer-aided design (CAD) model of a concrete implementation of the proposed mechanism, illustrating its key components and structural configuration. The primary actuator of the system, labelled as component A, is a linear actuator, which serves as the driving force for the knee joint. The B component represents a four-bar linkage system, which plays a crucial role in converting linear motion into controlled rotational movement. The C segment corresponds to the lower leg, specifically designed and adapted for future experimental validations. This lower-leg design guarantees that the mechanism's tip consistently goes to the same position whenever it stretches. As a result, this feature effectively functions as the mechanical end-stop of the component D, preventing excessive displacement (see Fig. \ref{Spring Opening}).
Additionally, the system incorporates a spring-based transmission mechanism, denoted as component D, which consists of uniformly structured springs. In the prototype that we designed for this study, each of these springs is strategically placed within the joint system and possesses a stiffness of 1.17  Nm/rad. The pre-load applied to each joint is determined using \eqref{preload}. In this formulation, ${k_{spring}}$ represents the torsional stiffness coefficient of the spring, while ${\delta}$ denotes the angular displacement of the spring relative to its neutral position. ${L}$ corresponds to the length of the torsional spring arm at which the force is exerted on the spring. This relationship is essential for accurately quantifying the pre-load imposed on the torsional spring. 

\begin{equation}
    F_{spring} = k_{spring}*\frac{\delta}{L}
    \label{preload}
\end{equation}

We ensure that the mechanism remains inactive (closed) with the pre-loaded springs until a specific force threshold is exceeded. In total, the system consists of six joints, with each joint being supported by four individual springs. Therefore, the total stiffness, 0.78 Nm/rad of the transmission mechanism ${k_{total}}$ is calculated from (\ref{total stiffness}). 
 \begin{equation}
   \frac{1}{k_{total}}= \sum^6_{i=1}\frac{1}{k_{joint_i}}
    \label{total stiffness}
\end{equation}

\section{Validation of the Transmission Mechanism}

To evaluate the effectiveness of the proposed LBVT mechanism, we conducted  2 simulation-based analyses. Each simulation compares two cases: one with the LBVT mechanism, where the transmission ratio changes after the activation threshold, and one without it, where the tip position is fixed in the closed condition of the LBVT and rigidly attached to the lower leg. In the first simulation, we investigated the torque amplification through the range of motion of the leg. The primary objective was to assess how the mechanism dynamically adjusts its transmission ratio in response to increasing torque demands and to compare its performance with a rigid transmission system without the LBVT mechanism. The second simulation investigates the triggering effect of the LBVT mechanism. Through these validations, we aim to confirm the torque amplification capability and investigate the threshold activation force required to trigger the mechanism.

\begin{figure} [ht]
    \centering
    \includegraphics[width=0.8\linewidth]{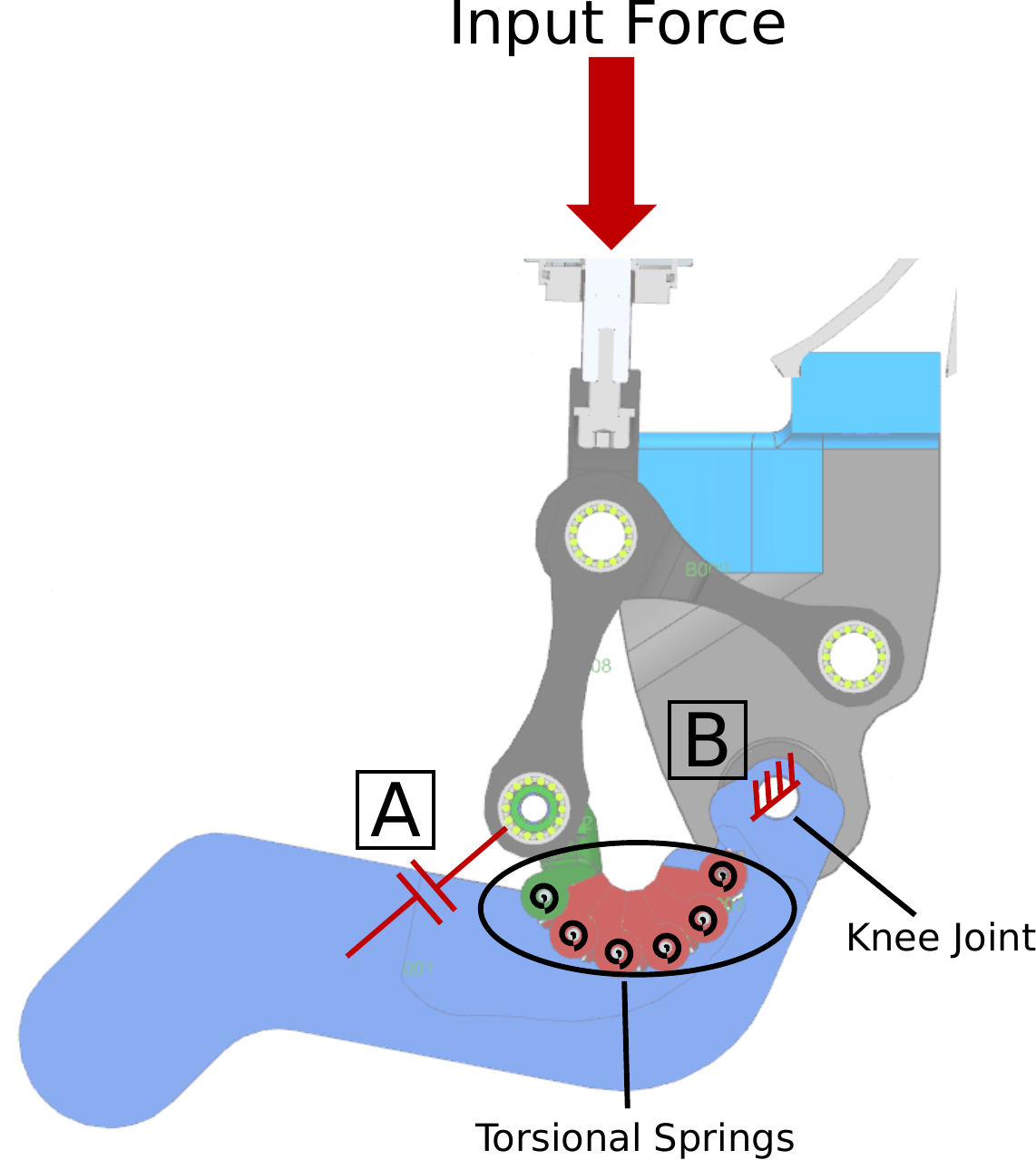}
    \caption{Simulation model of the transmission mechanism. A is the contact constraint between the lower leg and the LBVT mechanism. B is the fixed constraint on the knee joint.}
    \label{SimModel}
\end{figure}
\begin{figure*} [ht]
    \centering
    \includegraphics[width=1\linewidth,height=0.4\linewidth]{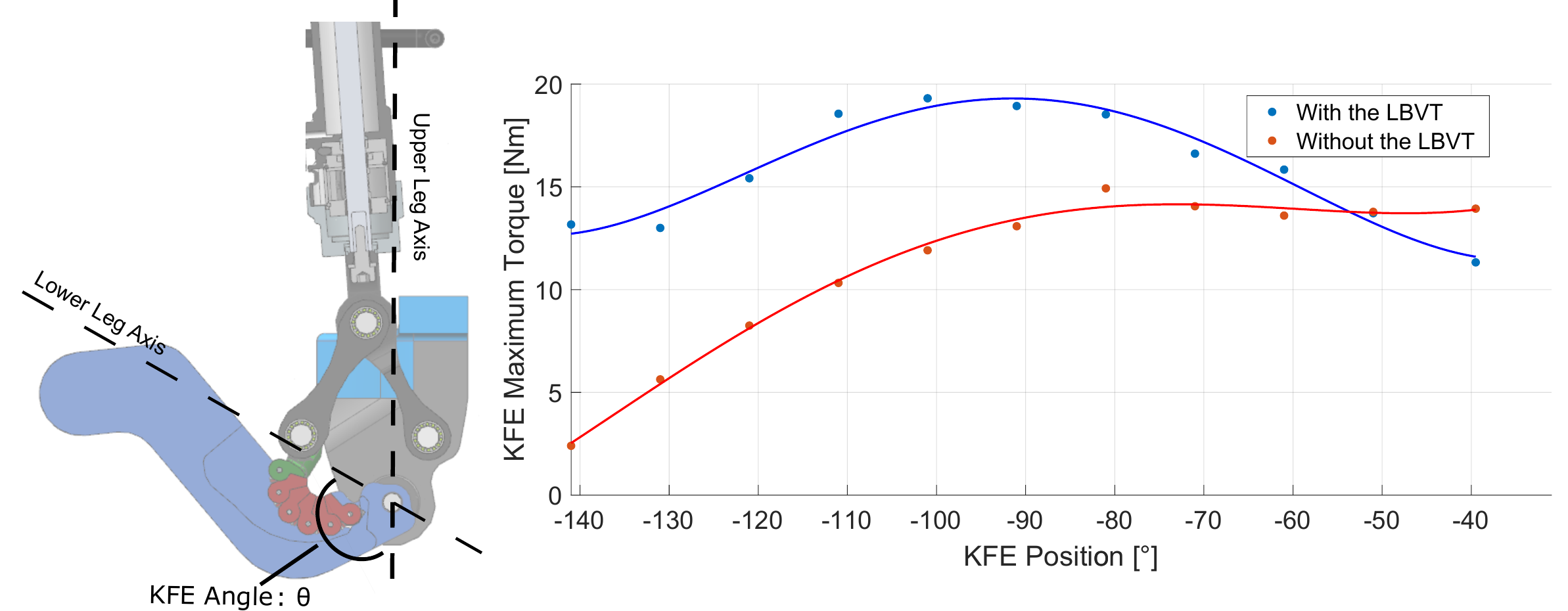}
    \caption{Comparison of the maximum torque profile of the KFE joint with and without the proposed LBVT mechanism. The KFE angle is defined as the angle between the centerlines of the upper leg and lower leg.}
    \label{Torque Profile}
\end{figure*}

This section shows an overview of the simulations along with their corresponding results. Each simulation is conducted using Siemens NX 9.0, a software tool for engineering design and analysis. For this study, we used the rigid body dynamics tool in the software. Fig. \ref{SimModel} depicts the general simulation model. To gain a precise understanding of the mechanism's behaviour, we established connections between the spring component and the lower leg. Four parallel torsional springs with a stiffness of 1.17  Nm/rad are attached at each joint of the spring component. An input force is exerted along the linear shaft of the actuator, ensuring accurate force transmission through the system. To accurately assess the torque output of the mechanism and streamline the simulation process, the knee joint is deliberately locked in place. This constraint eliminates unnecessary degrees of freedom, allowing for precise torque measurement and enhancing the reliability of the computational analysis.

The first simulation analyzes the maximum torque profile of the KFE joint throughout its range of motion. The corresponding torque profile is illustrated in Fig. \ref{Torque Profile}. Torque data is systematically recorded at KFE joint positions ranging from -141$^{\circ}$  to -39.5$^{\circ}$ in 10$^{\circ}$ increments for the configuration with and without the LBVT mechanism. The simulation is conducted individually for each increment (or joint position) for which the knee joint is constrained, and the data is interpolated using the polynomial interpolation method in MATLAB. It is observed that the implementation of the LBVT mechanism enhances the maximum torque output of the KFE joint. Also, we see the maximum torque enhanced at -88$^{\circ}$. However, beyond a flexion angle of -55$^{\circ}$, the amplification effect diminishes, leading to a progressive decrease in the maximum torque. 

The goal of the second simulation was to gain deeper insights into the mechanism's behaviour under specific loading conditions. In this simulation, the analysis was constrained to a single KFE joint position at -88$^{\circ}$ to isolate its response as mentioned earlier. This position was selected to clearly highlight the torque amplification effect. The linear actuator was applying a 165~N force to the four-bar linkage, and the corresponding torque output at the KFE joint was recorded. Additionally, the displacement variations within the LBVT mechanism were monitored to assess its dynamic response. This focused investigation provides a clearer understanding of the torque transmission characteristics and the mechanical interaction between the LBVT mechanism and the KFE joint, offering valuable insights for optimizing the system’s performance.

\begin{figure} [ht]
    \centering
    \includegraphics[width=1\linewidth]{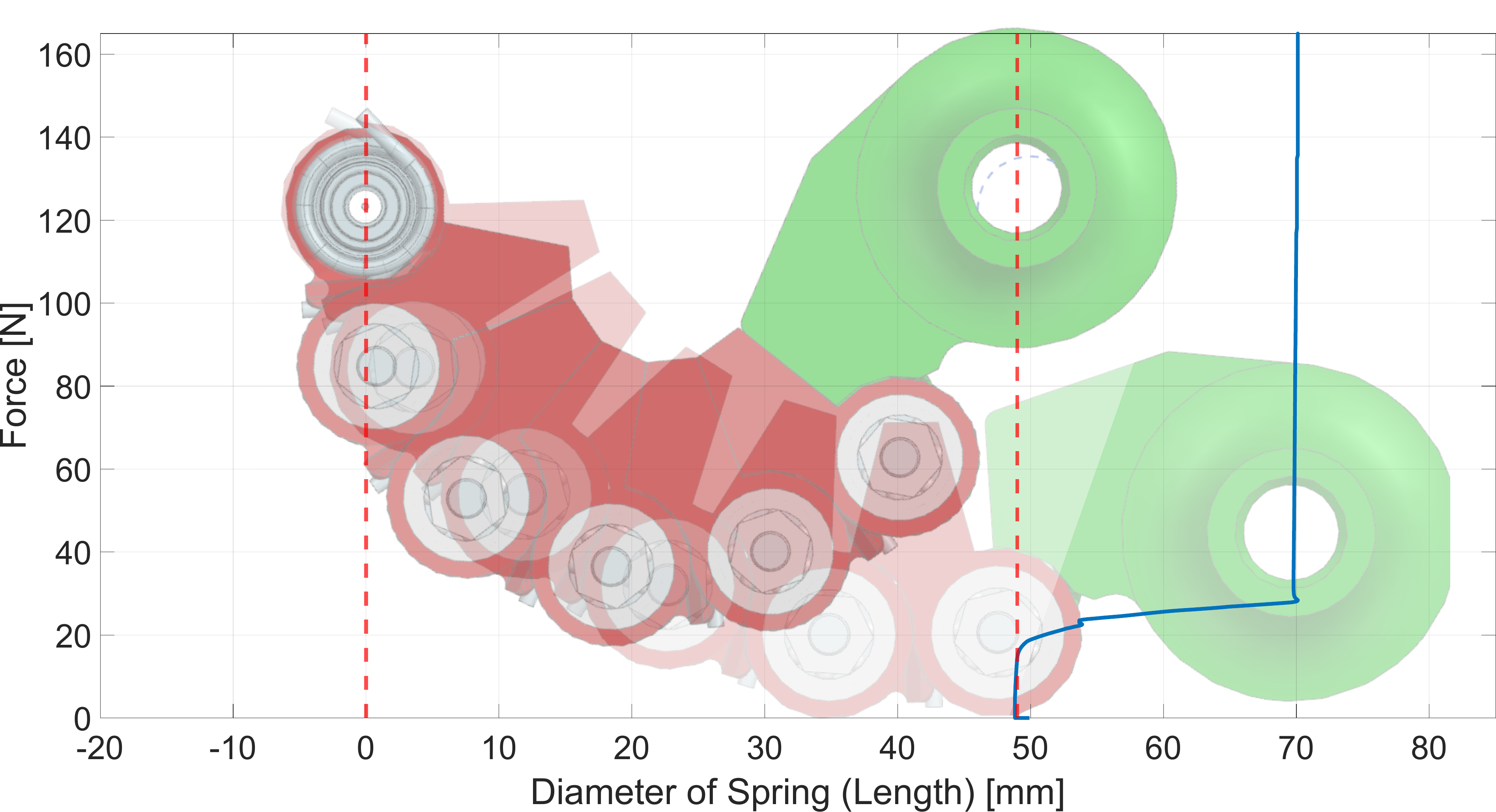}
    \caption{Triggering effect of the transmission mechanism. The x-axis shows the diameter of the spring component, which corresponds to the distance between the anchor point and the tip of the LBVT mechanism (see Fig. \ref{Kinmeatics}).}
    \label{Trigger}
\end{figure}

To evaluate the triggering effect of the mechanism, the displacement response was analyzed in relation to the applied input force. This analysis allows for the identification of the force threshold at which the mechanism gets activated. Fig. \ref{Trigger} illustrates the triggering force of the system, where the spring element diameter represents the distance between the anchor point of the spring element on the lower leg and the tip of the LBVT mechanism. As depicted in the plot, a noticeable change in the spring element diameter begins to occur beyond an applied force of 18~N. Consequently, the threshold force required to activate the mechanism is determined to be 20 N on the linear actuator side. This value was selected for use in upcoming experimental validations planned with the existing hardware. 

\begin{figure} [ht]
    \centering
    \includegraphics[width=0.9\linewidth]{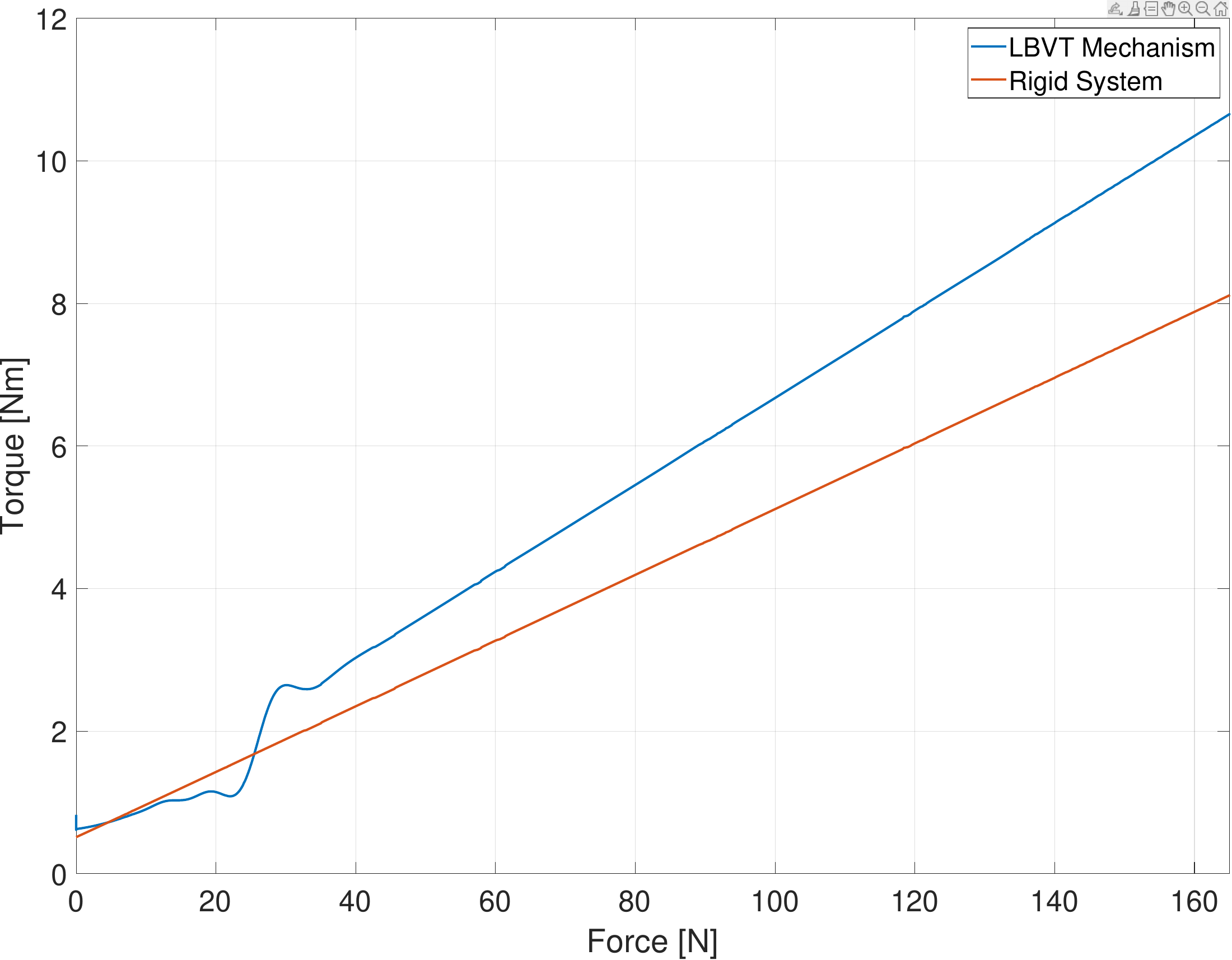}
    \caption{Comparison between the LBVT  mechanism and the rigid system through simulation. The rigid system represents the four-bar linkage without the spring component. The blue line represents the proposed LBVT mechanism. The red line represents the system without the LBVT mechanism, which means it is a rigid system.}
    \label{TorqueForce}
\end{figure}

As the KFE torque amplification is influenced by variations in the spring diameter, it is inherently dependent on the force input from the linear actuator. To analyze this relationship, a comparative study was conducted between two configurations: one incorporating the LBVT mechanism and the other without it. The investigation focused on evaluating the variations in KFE torque concerning changes in the applied input force. Fig. \ref{TorqueForce} presents the KFE torque as a function of the input force for both configurations. As observed, the slope of the curve representing the system with the LBVT mechanism increases above an input force of 18~N. This indicates that once the applied force surpasses the threshold, the output torque experiences a progressive amplification. This behaviour highlights the nonlinear response of the mechanism, demonstrating its ability to enhance torque transmission above a critical activation point.

\begin{figure} [ht]
    \centering
    \includegraphics[width=1\linewidth]{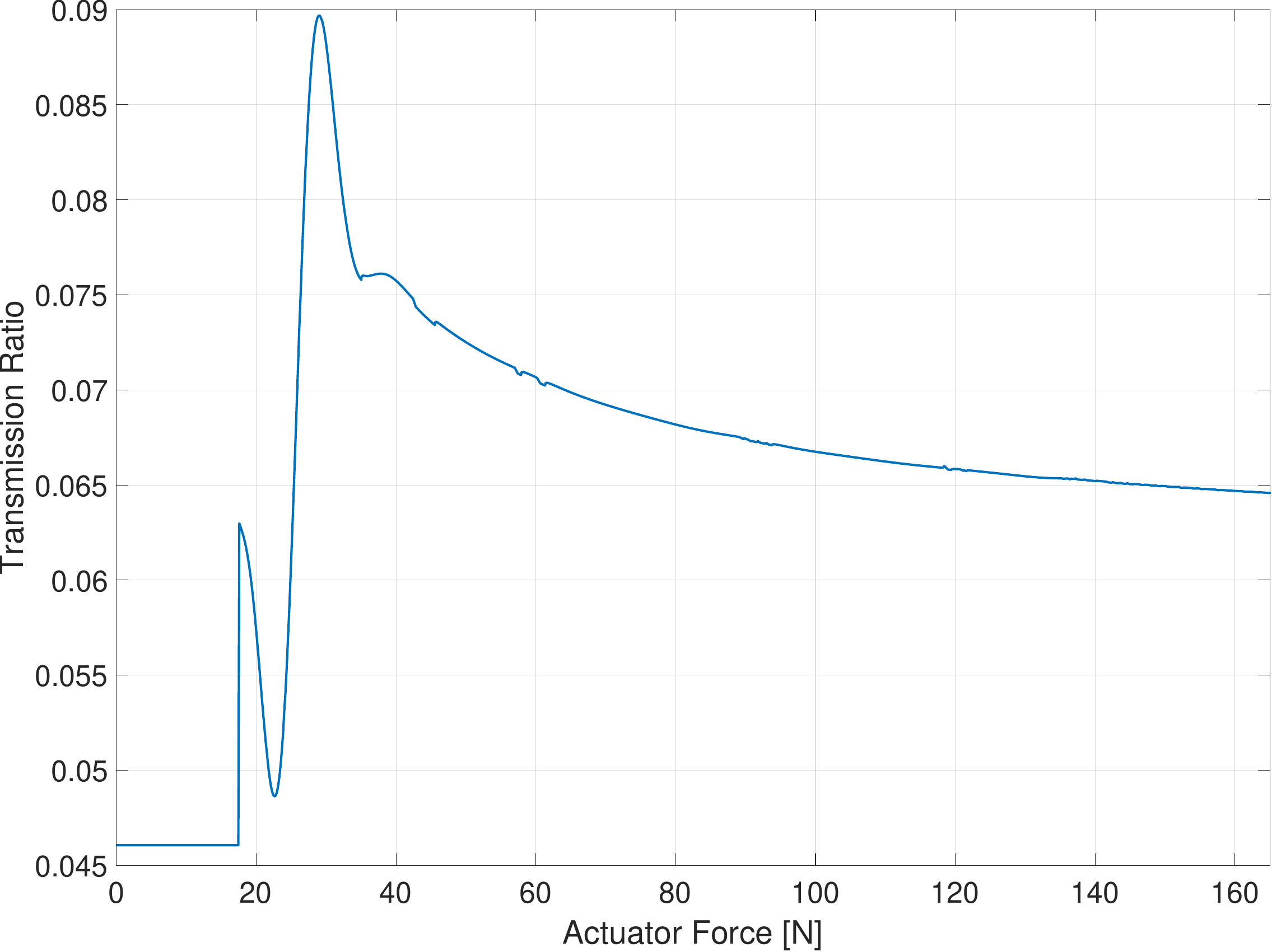}
    \caption{Transmission ratio w.r.t input force. The curve's peak results from the impact force and the contact modelling used in the simulation. This behaviour is observed because of the contact dynamics defined by the material and contact properties in the software. A similar short-duration response may occur in real hardware due to localized deformation and impact between components.}
    \label{Gear}
\end{figure}

Subsequently, the analysis was extended to examine the variable transmission ratio of the LBVT mechanism. Fig. \ref{Gear} illustrates the variation in transmission ratio as a function of the applied input force. The results indicate that once the input force surpasses the threshold value, the transmission ratio increases by 40$\%$, demonstrating the adaptive nature of the system. Furthermore, this analysis provides additional evidence of the mechanism's triggering behaviour, highlighting its capability to dynamically adjust the transmission ratio in response to external force variations.

\section{Discussion}

The simulation results presented in this study demonstrate the effectiveness of the LBVT mechanism. As expected, the torque amplification behaviour was observed to be dependent on the applied force. We identified a threshold activation force of 18~N for our prototype. Above the threshold, the system exhibited an increase in transmission ratio up to 40$\%$. This confirms the passive adaptability of the LBVT mechanism and its capability to dynamically adjust to external loads without the need for additional actuation.

The comparative analysis between the rigid transmission system and the LBVT system highlights the advantages of the proposed mechanism. The LBVT system demonstrated the torque amplification above the threshold force.

Despite the promising results, there are certain limitations of the current implementation. The current design has a limited operational range of 86$^{\circ}$ (-141$^{\circ}$  to -55$^{\circ}$) on the KFE joint position; this could affect the mechanism's adaptability in applications requiring larger joint positions. However, this range is still sufficient for tasks such as stance position or locomotion, where the knee joint primarily operates within this region. In future iterations, the range could be extended by adjusting the four-bar linkage design or modifying the anchor position of the elastic element.

The material properties and stiffness of the pre-tensioned springs play a crucial role in defining the force threshold and overall responsiveness of the mechanism. The current design uses commercial springs with a relatively low stiffness value, which may not be optimal for high-load conditions or full-scale robotic applications. This constraint results from the limited available space within the lower-leg structure, restricting the use of stiffer or larger spring components.

The results show that the LBVT increases the transmission ratio once the activation threshold is exceeded. Unlike active VTMs, this behavior is achieved without a secondary actuator, avoiding additional weight and complexity. Compared to conventional passive systems, which typically lack effective real-time adaptation, the LBVT demonstrates torque amplification and autonomous adjustment under load. In this way, it combines the simplicity of passive designs with adaptability that approaches active mechanisms. In light of these results, the LBVT can be considered for applications where compact design, low complexity, and adaptive torque output are critical, such as legged robots operating in unstructured environments.

\section{Conclusion}

In this study, we introduced a Load-Based Variable Transmission mechanism that is designed to enhance robotic actuation by passively adjusting its transmission ratio in response to joint torque demands. Unlike conventional variable transmission mechanisms that require active control through a second actuator, our proposed system leverages pre-tensioned springs to enable autonomous adaptation. Our design removes the need for additional actuators, reducing the complexity of the system while improving torque delivery under varying load conditions.

Through simulations, we demonstrated that the LVBT mechanism effectively amplifies torque above a specific threshold input force. Our results show that the transmission ratio increases up to 40$\%$ once the activation force is reached, highlighting the system's capabilities. 

Overall, the LVBT mechanism presents a promising approach to passive variable transmission, offering a lightweight, efficient, and adaptable solution for robotic applications requiring torque amplification without additional actuation complexity. Despite its advantages, the current implementation has limitations, such as a constrained operational range and the dependency on spring stiffness and actuator workspace. For future work, we will focus on optimizing the mechanism's structural design to address these limitations and enhance its integration within multi-joint robotic systems such as quadruped robots. In addition, we plan to manufacture the first prototype of the proposed mechanism and conduct experiments on real hardware.

\bibliographystyle{IEEEtran}

\bibliography{references}

\begin{thebibliography}{10}
\providecommand{\url}[1]{#1}
\csname url@samestyle\endcsname
\providecommand{\newblock}{\relax}
\providecommand{\bibinfo}[2]{#2}
\providecommand{\BIBentrySTDinterwordspacing}{\spaceskip=0pt\relax}
\providecommand{\BIBentryALTinterwordstretchfactor}{4}
\providecommand{\BIBentryALTinterwordspacing}{\spaceskip=\fontdimen2\font plus
\BIBentryALTinterwordstretchfactor\fontdimen3\font minus \fontdimen4\font\relax}
\providecommand{\BIBforeignlanguage}[2]{{%
\expandafter\ifx\csname l@#1\endcsname\relax
\typeout{** WARNING: IEEEtran.bst: No hyphenation pattern has been}%
\typeout{** loaded for the language `#1'. Using the pattern for}%
\typeout{** the default language instead.}%
\else
\language=\csname l@#1\endcsname
\fi
#2}}
\providecommand{\BIBdecl}{\relax}
\BIBdecl

\bibitem{hirose1999development}
S.~Hirose, C.~Tibbetts, and T.~Hagiwara, ``Development of x-screw: a load-sensitive actuator incorporating a variable transmission,'' in \emph{Proceedings 1999 IEEE International Conference on Robotics and Automation (Cat. No. 99CH36288C)}, vol.~1.\hskip 1em plus 0.5em minus 0.4em\relax IEEE, 1999, pp. 193--199.

\bibitem{yamada2012radial}
H.~Yamada, ``A radial crank-type continuously variable transmission driven by two ball screws,'' in \emph{2012 IEEE International Conference on Robotics and Automation}.\hskip 1em plus 0.5em minus 0.4em\relax IEEE, 2012, pp. 1982--1987.

\bibitem{lee2011new}
H.~Lee and Y.~Choi, ``A new actuator system using dual-motors and a planetary gear,'' \emph{IEEE/ASME Transactions on mechatronics}, vol.~17, no.~1, pp. 192--197, 2011.

\bibitem{kim2007double}
B.-S. Kim, J.-J. Park, and J.-B. Song, ``Double actuator unit with planetary gear train for a safe manipulator,'' in \emph{Proceedings 2007 IEEE International Conference on Robotics and Automation}.\hskip 1em plus 0.5em minus 0.4em\relax IEEE, 2007, pp. 1146--1151.

\bibitem{girard2015two}
A.~Girard and H.~H. Asada, ``A two-speed actuator for robotics with fast seamless gear shifting,'' in \emph{2015 IEEE/RSJ International Conference on Intelligent Robots and Systems (IROS)}.\hskip 1em plus 0.5em minus 0.4em\relax IEEE, 2015, pp. 4704--4711.

\bibitem{girard2017leveraging}
A.~Girard and H.~Asada, ``Leveraging natural load dynamics with variable gear-ratio actuators,'' \emph{IEEE Robotics and Automation Letters}, vol.~2, no.~2, pp. 741--748, 2017.

\bibitem{dunwen2011concept}
W.~Dunwen, G.~Wenjie, and L.~Yiyang, ``The concept of a jumping rescue robot with variable transmission mechanism,'' in \emph{2011 IEEE International Symposium on Safety, Security, and Rescue Robotics}.\hskip 1em plus 0.5em minus 0.4em\relax IEEE, 2011, pp. 99--104.

\bibitem{sun2018variable}
X.~Sun, F.~Sugai, K.~Okada, and M.~Inaba, ``Variable transmission series elastic actuator for robotic prosthesis,'' in \emph{2018 IEEE International Conference on Robotics and Automation (ICRA)}.\hskip 1em plus 0.5em minus 0.4em\relax IEEE, 2018, pp. 2796--2803.

\bibitem{girard2022fast}
A.~Girard and H.~H. Asada, ``A fast gear-shifting actuator for robotic tasks with contacts,'' \emph{arXiv preprint arXiv:2205.15137}, 2022.

\bibitem{meng2022explosive}
F.~Meng, Q.~Huang, Z.~Yu, X.~Chen, X.~Fan, W.~Zhang, and A.~Ming, ``Explosive electric actuator and control for legged robots,'' \emph{Engineering}, vol.~12, pp. 39--47, 2022.

\bibitem{song2022drpd}
T.-G. Song, Y.-H. Shin, S.~Hong, H.~C. Choi, J.-H. Kim, and H.-W. Park, ``Drpd, dual reduction ratio planetary drive for articulated robot actuators,'' in \emph{2022 IEEE/RSJ International Conference on Intelligent Robots and Systems (IROS)}.\hskip 1em plus 0.5em minus 0.4em\relax IEEE, 2022, pp. 443--450.

\bibitem{wang2024development}
T.~Wang, H.~Wen, Z.~Song, Z.~Dong, and C.~Liu, ``Development of variable transmission series elastic actuator for hip exoskeletons,'' in \emph{2024 IEEE International Conference on Robotics and Automation (ICRA)}.\hskip 1em plus 0.5em minus 0.4em\relax IEEE, 2024, pp. 7055--7061.

\bibitem{kim2020elastomeric}
S.~Kim, J.~Sim, and J.~Park, ``Elastomeric continuously variable transmission combined with twisted string actuator,'' \emph{IEEE Robotics and Automation Letters}, vol.~5, no.~4, pp. 5477--5484, 2020.

\bibitem{shin2022passively}
W.~Shin, S.~Park, G.~Park, and J.~Kim, ``A passively adaptable toroidal continuously variable transmission combined with twisted string actuator,'' in \emph{2022 International Conference on Robotics and Automation (ICRA)}.\hskip 1em plus 0.5em minus 0.4em\relax IEEE, 2022, pp. 11\,409--11\,415.

\bibitem{lee2022mechanical}
E.~Lee, H.~Song, J.~Jeong, and S.~Jeong, ``Mechanical variable magnetic gear transmission: Concept and preliminary research,'' \emph{IEEE Robotics and Automation Letters}, vol.~7, no.~2, pp. 3357--3364, 2022.

\bibitem{nobaveh2023compliant}
A.~A. Nobaveh, J.~L. Herder, and G.~Radaelli, ``A compliant continuously variable transmission (cvt),'' \emph{Mechanism and Machine Theory}, vol. 184, p. 105281, 2023.

\bibitem{kim2020optimization}
J.~H. Kim and I.~G. Jang, ``Optimization-based investigation of bioinspired variable gearing of the distributed actuation mechanism to maximize velocity and force,'' \emph{IEEE Robotics and Automation Letters}, vol.~5, no.~4, pp. 6326--6333, 2020.

\bibitem{sanz2023active}
C.~B. Sanz-Mor{\`e}re, M.~Fantozzi, F.~dell'Agnello, A.~Baldoni, F.~Giovacchini, S.~Crea, and N.~Vitiello, ``An active knee orthosis with a variable transmission ratio through a motorized dual clutch,'' \emph{Mechatronics}, vol.~94, p. 103018, 2023.

\bibitem{valsecchi2023barry}
G.~Valsecchi, N.~Rudin, L.~Nachtigall, K.~Mayer, F.~Tischhauser, and M.~Hutter, ``Barry: a high-payload and agile quadruped robot,'' \emph{IEEE Robotics and Automation Letters}, 2023.

\bibitem{valsecchi2023actively}
G.~Valsecchi, F.~Tischhauser, J.~Junger, Y.~Bernarnd, and M.~Hutter, ``Actively variable transmission robotic leg,'' in \emph{Climbing and Walking Robots Conference}.\hskip 1em plus 0.5em minus 0.4em\relax Springer, 2023, pp. 40--51.

\bibitem{shin2025variable}
W.~Shin, S.~Park, S.~Kwon, B.~Ahn, and J.~Kim, ``Variable transmission with actively controllable reduction ratio,'' \emph{IEEE/ASME Transactions on Mechatronics}, 2025.

\bibitem{hur2024continuously}
J.~Hur, H.~Song, and S.~Jeong, ``Continuously variable transmission and stiffness actuator based on actively variable four-bar linkage for highly dynamic robot systems,'' \emph{IEEE Robotics and Automation Letters}, 2024.

\bibitem{semini2016design}
C.~Semini, V.~Barasuol, J.~Goldsmith, M.~Frigerio, M.~Focchi, Y.~Gao, and D.~G. Caldwell, ``Design of the hydraulically actuated, torque-controlled quadruped robot hyq2max,'' \emph{IEEE/Asme Transactions on Mechatronics}, vol.~22, no.~2, pp. 635--646, 2016.

\end{thebibliography}
\end{document}